\documentclass[journal]{IEEEtran}

\usepackage{cite}
\usepackage{amsmath,amssymb,amsfonts}
\usepackage{graphicx}
\graphicspath{{images/}}
\usepackage[dvipsnames]{xcolor}
\usepackage{soul}
\usepackage{url}
\usepackage{multirow}
\usepackage{array,booktabs}
\usepackage{xspace}
\usepackage[colorinlistoftodos]{todonotes}
\usepackage[normalem]{ulem}
\usepackage{hyperref}

\usepackage{siunitx}
\usepackage{algorithm}
\usepackage{algpseudocode}
\usepackage{fixltx2e}
\usepackage{tikz}
\usepackage{adjustbox}
\usepackage{tabstackengine}

\ifCLASSOPTIONcompsoc
  \usepackage[caption=false,font=normalsize,labelfont=sf,textfont=sf]{subfig}
\else
  \usepackage[caption=false,font=footnotesize]{subfig}
\fi

\newcommand{\davinci}{da~Vinci\textsuperscript\textregistered\xspace}

\begin{document}
%
\title{Improving rigid 3D calibration for robotic surgery}

\author{Andrea Roberti, Nicola Piccinelli, Daniele Meli, Riccardo Muradore, Paolo Fiorini
\thanks{Authors are with the Department of Computer Science, University of Verona, strada Le Grazie 15, 37135 Verona, Italy.}
\thanks{Contact author: Andrea Roberti, \texttt{andrea.roberti@univr.it}}
\thanks{Manuscript received XXX; revised YYY.}}

\markboth{IEEE Transactions on Medical Robotics and Bionics}%
{Shell \MakeLowercase{\textit{et al.}}: Bare Demo of IEEEtran.cls for IEEE Journals}

\maketitle

\begin{abstract}
Autonomy is the frontier of research in robotic surgery and its aim is to improve the quality of surgical procedures in the next future. One fundamental requirement for autonomy is advanced perception capability through vision sensors. In this paper, we propose a novel calibration technique for a surgical scenario with \davinci robot. Calibration of the camera and the robot is necessary for precise positioning of the tools in order to emulate the high performance surgeons. Our calibration technique is tailored for RGB-D camera. Different tests performed on relevant use cases for surgery prove that we significantly improve precision and accuracy with respect to the state of the art solutions for similar devices on a surgical-size setup. Moreover, our calibration method can be easily extended to standard surgical endoscope to prompt its use in real surgical scenario.
\end{abstract}



%
\IEEEpeerreviewmaketitle

\section{Introduction}
A significant part of the research in Robotic-assisted Minimally Invasive Surgery (R-MIS) is nowday focussing on the development of supporting autonomous systems for the execution of repetitive surgical steps, such as suturing, ablation and microscopic image scanning~\cite{du2016combined}. 
This helps the surgeon, since she/he can focus on more complex cognitive phases of the procedure, while interacting with the robot performing low-level maneuvers.
Autonomy requires systems with advanced capabilities in perception, reasoning and motion planning, as demostrated in~\cite{ARS2019,DeRossi2019a}.
Specifically, advances in medical imaging and vision techniques have significantly improved the performance of robotic surgical systems in a range of clinical scenarios, such as orthopaedics and neurosurgery~\cite{adebar20143}. Vision systems can retrieve pre and intra operative information from tomography (CT)~\cite{ISMR2020_altair}, magnetic resonance (MR)
and ultrasound to guide trajectory execution and support the surgeon in decision making. However, in order to perform image-guided interventions, an accurate calibration is necessary to represent poses of robots, instruments and anatomy in a common reference frame. 
In this paper, we address the problem of accurate calibration in a surgical setup, using the \davinci Research Kit (dVRK). We present our preliminary setup with a RGB-D camera, and we perform exhaustive experimental validation on relevant use cases for surgery.
We separate the calibration of the two slave manipulators from the hand-eye calibration of the camera. For both calibrations we propose a three steps method with closed-form solution:
\begin{itemize}
    \item reaching reference points on a custom calibration board with the end-effector of the surgical robot.
    \item recognizing the marker and the points in 3D space with the camera. 
    \item mapping them with the poses reached by the slave manipulators in the previous step.
\end{itemize}
The main advantage of the proposed method is the improved accuracy in a 3D metric space, which is increased by four times with respect to the state-of-the-art results with comparable sensors \cite{table}. Moreover, with our method the camera can be mounted on the moving endoscopic arm of the \davinci, overcoming the limitations of a fixed camera.

This paper is organized as follows. In Section \ref{sec:sota} we review the main approaches to 3D rigid calibration with a RGB-D camera. In Section \ref{sec:methods} we describe our calibration technique and in Section \ref{sec:exp} we validate the proposed method performing grasping and dual-arm manipulation on the dVRK. Section \ref{sec:conclusion} concludes the paper and outlines possible future research.

\section{Related works}\label{sec:sota}
Hand-eye calibration has been widely studied within the robotics literature~\cite{shah2012overview}. Thus far, several closed-form solutions for 2d images have been proposed for hand-eye calibration that solve the problem using linear methods by separating rotation and translation~\cite{shiu1989calibration}. The orientation component was derived by utilizing the angle-axis formulation of rotation, while the translational component could be solved using standard linear systems techniques once the rotation part is estimated. Chou and 
Kamel~\cite{chou1991finding} introduced quaternion to represent orientation and solved the rotation group as a homogeneous linear least squares problem. Closed form solution was then derived using the generalized inverse method with singular value decomposition analysis. Other works~\cite{liang2008hand, pan2014closed, najafi2014closed} used the Kronecker product to get the homogeneous linear equation for the rotation matrix.
However, all these methods separate the orientation component from the translational one, while the errors on the two components are correlated. 
Working directly on the 3D space avoids this kind of errors. In~\cite{table} the authors studied the comparison between hand-eye calibration based on 2D and 3D images, introducing quantitative 2D and 3D error metrics for the accuracy of the calibration. They proved that the 3D calibration approach provides more accurate results on average but requires burdensome manual preparation and much more computation time than the 2D approaches. Kim used 3D measurements at the center of markers for the hand-eye calibration \cite{kim2013hand}. Fuchs~\cite{fuchs2012calibration} proposed a solution which uses depth measurements, instead of 2D images, using a calibration plane with known position and orientation. The hand-eye calibration is then estimated by solving a least squares curve fitting problem of the measured depth values with the calibration plane. With our method the calibration board can be positioned without any prior knowledge without changing the accuracy and computational time within the work area. 

\section{Materials and methods}\label{sec:methods}
\begin{figure}[b]
    \centering
    \includegraphics[width=\columnwidth]{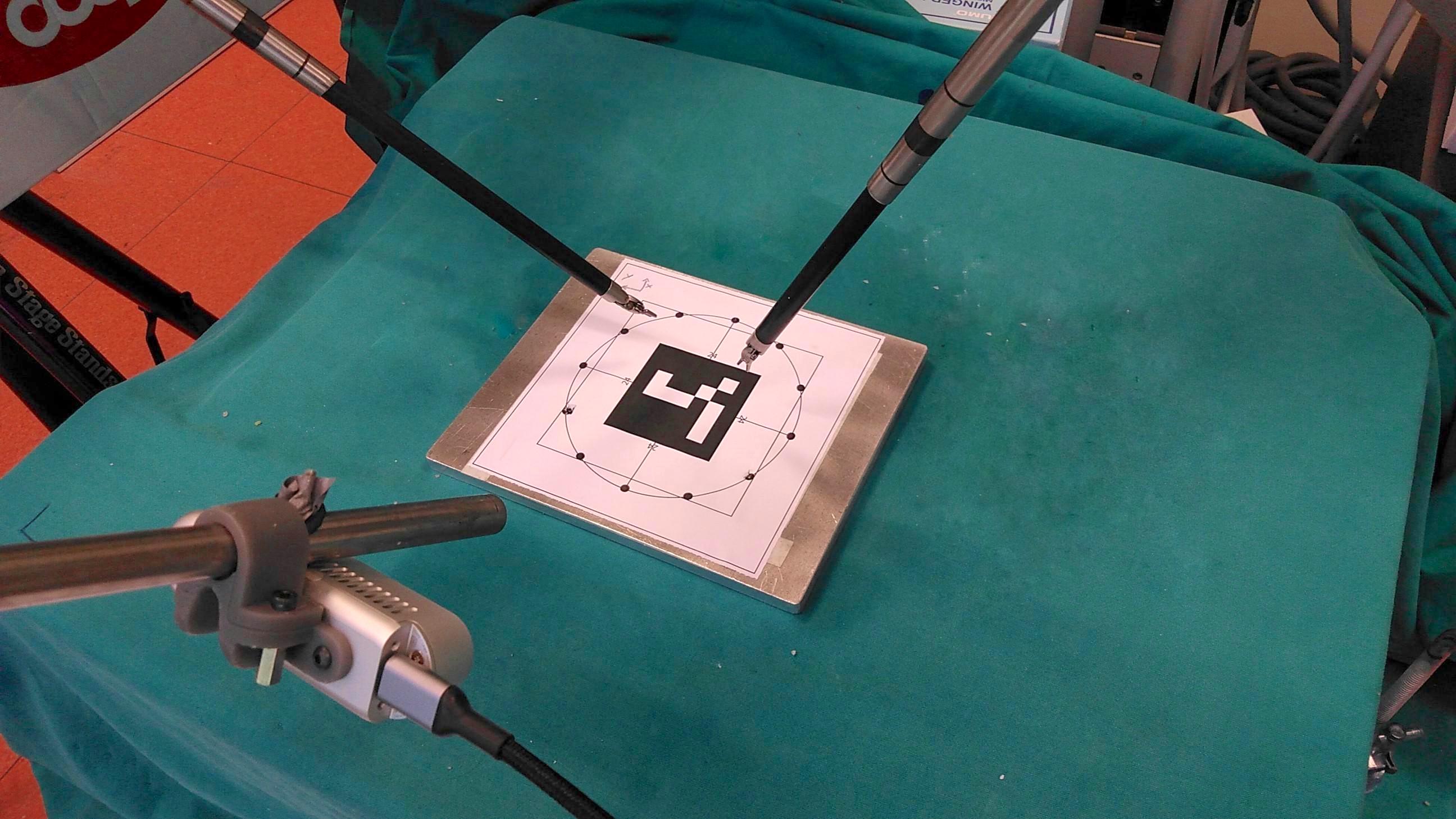}
    \caption{The proposed calibration system. The RealSense d435, the PSMs and the calibration pattern.}
    \label{fig:adapterRS}
\end{figure}

\begin{figure}[t]
    \centering
    \subfloat[Calibration pattern]{
        \begin{adjustbox}{width=0.5\columnwidth}
        \begin{tikzpicture}
            \node[inner sep=0pt] at (0,0) {\includegraphics[width=\columnwidth,angle=-180,origin=c]{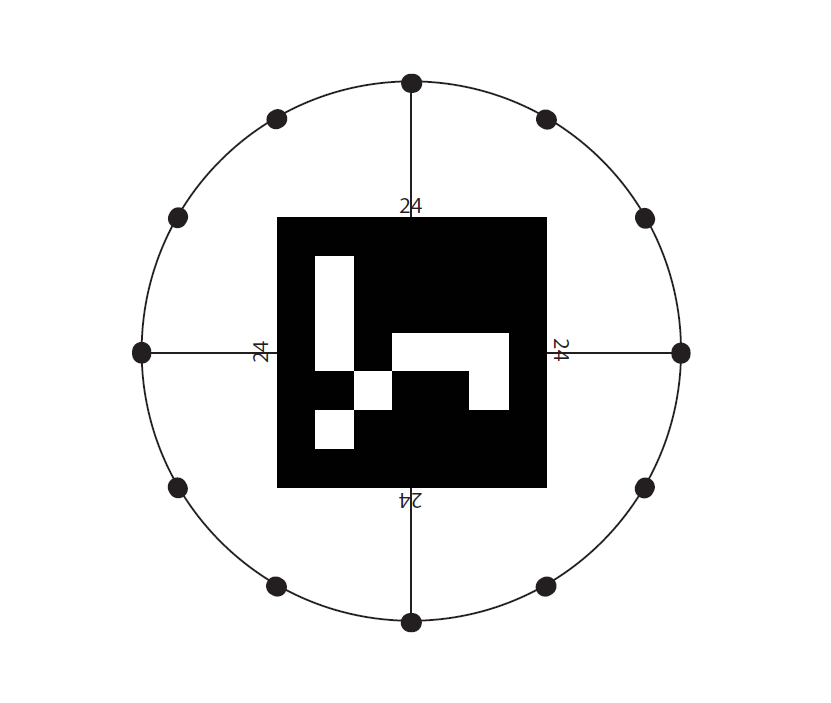}};

            \draw[->,red] (-2.5,-3.5) -- (-2.5,-2.5);
            \node at (-2.75,-2.5) {$x$};
            \draw[->,green] (-2.5,-3.5) -- (-3.5,-3.5);
            \node at (-3.5,-3.25) {$y$};
            \draw[fill,blue] (-2.5,-3.5) circle (2pt);
            \node at (-2.25,-3.75) {$z$};
        \end{tikzpicture}
        \end{adjustbox}
        \label{fig:marker}
    }
    \subfloat[ECM adapter
    ]{\includegraphics[width=0.5\columnwidth]{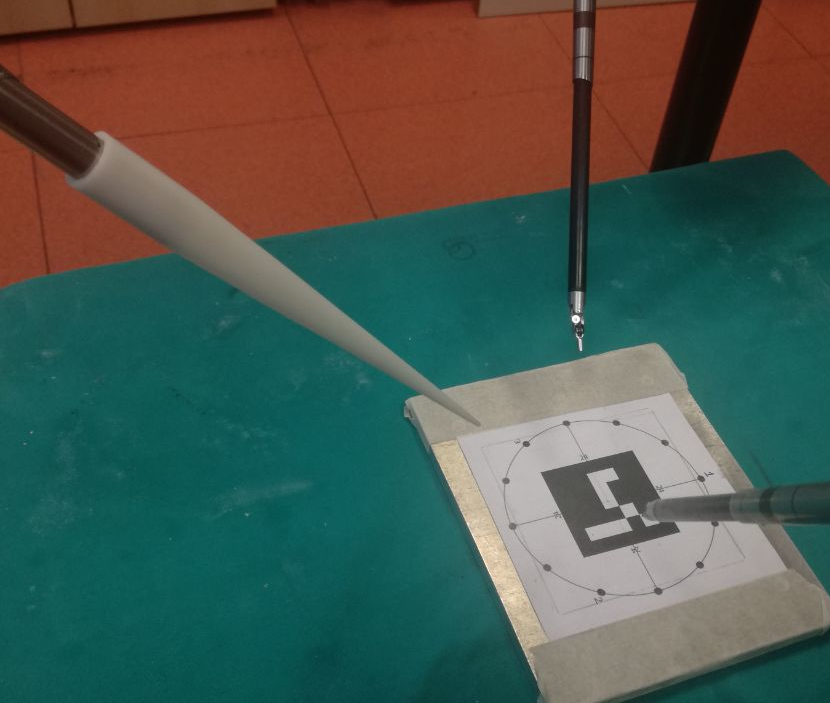}\label{fig:adapterECM}}\caption{The calibration components. a) the calibration board with the marker, the coloured axes represents the common reference frame directions b) the adapter for the ECM positioning.}
    \label{fig:calibration}
\end{figure}
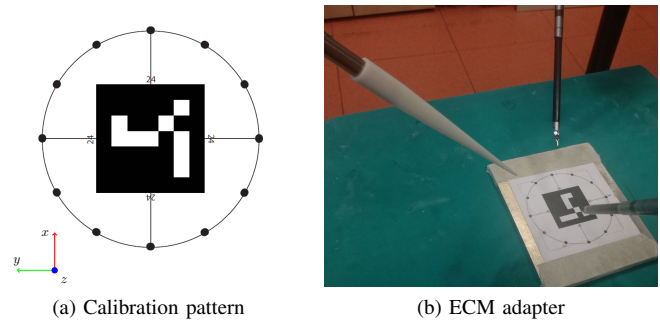

The setup used for the registration between all the frames involved in a surgical task is shown in Figure~\ref{fig:adapterRS}.
It consists of the \davinci surgical robot controlled through the \davinci Research Kit (dVRK) and an Intel RealSense d435 RGB-D camera. We use a custom calibration board (Figure~\ref{fig:marker}) with an Aruco marker in the center of a circumference of 5 \si{\cm} radius, with several reference dots $D$ on it. We also rigidly mount the RGB-D on the \davinci Endoscope Camera Manipulator (ECM) with a 3D-printed adapter (Figure~\ref{fig:adapterECM}).

The procedure starts by positioning the calibration board in the robot workspace, in such a way that reference dots can be reached by the Patient Side Manipulators (PSMs) and the ECM. We choose a subset of reference points $P \subset D$ such that each point $p \in P$ is reachable by the arms. The points in $P$ must be symmetric with respect to the center of the board to compute the origin of the common reference frame, and at least three points are needed to estimate the plane coefficients. We execute Algorithm \ref{alg:GFR} to generate a common reference plane and frame for all the tools. First, we use tele-operation provided by the dVRK to position the end effector of the PSMs and the ECM on points in $P$. To this aim, we mount the 3D-printed adapter shown in Figure \ref{fig:adapterECM} on the ECM. We touch points with each arm in the same order, then we place the tools above the board to define the plane normal direction. 

Afterwards, we proceed with the 3D hand-eye calibration of the camera.
We first detect the center of the Aruco marker on the board with respect to the camera frame. Once we get a stable pose of the marker, we align the pose on the point cloud generated from the depth map. 

Finally, we use the marker pose and its known radius to generate the pose of every dot in the set $P$ in the marker reference frame, as well as the point above the calibration board.
Given the poses of every point $p \in P$ with all the tools and the camera, we estimate a plane applying Random Sample Consensus (RANSAC) \cite{fischler1981random}. Hence, the centroid of all points on the plane is chosen as the origin of the common reference frame for the tools.

\begin{algorithm}[t]
    \caption{Multi arm reference frame calibration}\label{alg:GFR}
    \begin{algorithmic}[1]
        \State \textbf{Input}: number $n$ of Points $p \in Q = P \cup \{point_{top}\}$; Tools; 
        \State \textbf{Output}: Rigid trasformation between all the frames
        \For{ t = 1 to \textit{n}}
            \If {Tool = camera}
                \State \textit{Aruco marker identification}
                \State \textit{Identification of $p[t]$  $\gets$ Point Cloud $P_{in}$}
                \State \Return {$ Pose_{p}[n]$}
            \ElsIf {Tool = robotic arm}
                    \State \textit{Forward kinematic to $p[t]$ }
                    \State \Return {$ Pose_{p}[n]$}
            \EndIf
            \For{ t = 1 to \textit{$ Pose_{p}[n]$}}
                \State \textit{Plane Estimation $\gets$ RANSAC}
            \EndFor
            
        \EndFor
        
    \end{algorithmic}
\end{algorithm}

\section{Experiments}\label{sec:exp}
\begin{figure}[t]
    \centering
    \includegraphics[height=0.6\columnwidth]{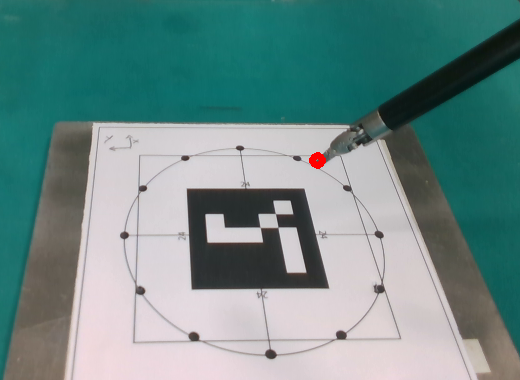}
    \caption{An example of re-projection of \davinci surgical instruments by using kinematic re-projection of the model directly onto camera color image.}
    \label{fig:calibrationRes}
\end{figure}

Figure~\ref{fig:calibrationRes} shows the re-projection of one \davinci surgical instrument in the camera image plane. The calibration pipeline has been implemented using Robot Operating System (ROS) to handle all the frames involved during the workflow.
Data from the camera are processed using the Point Cloud Library (PCL) and OpenCV. Since the calibration error depends on the estimation of the plane orientation, we first test the accuracy of our method moving both PSMs to the same poses 5 \si{\cm} above the calibration board, and we remain within the external circumference of the calibration board. Then, we measure the Euclidean distance between the poses of the PSMs. The maximum error is less than 0.8 \si{\mm} within 5 \si{\cm} far from the origin of the common reference frame. This error is lower than the 4-\si{\mm} minimum error obtained in \cite{table, realsense_kuka} with comparable camera hardware and on larger scale.

\begin{figure}[t]
    \centering
    \subfloat[Localization and grasping]{\includegraphics[height=0.35\columnwidth]{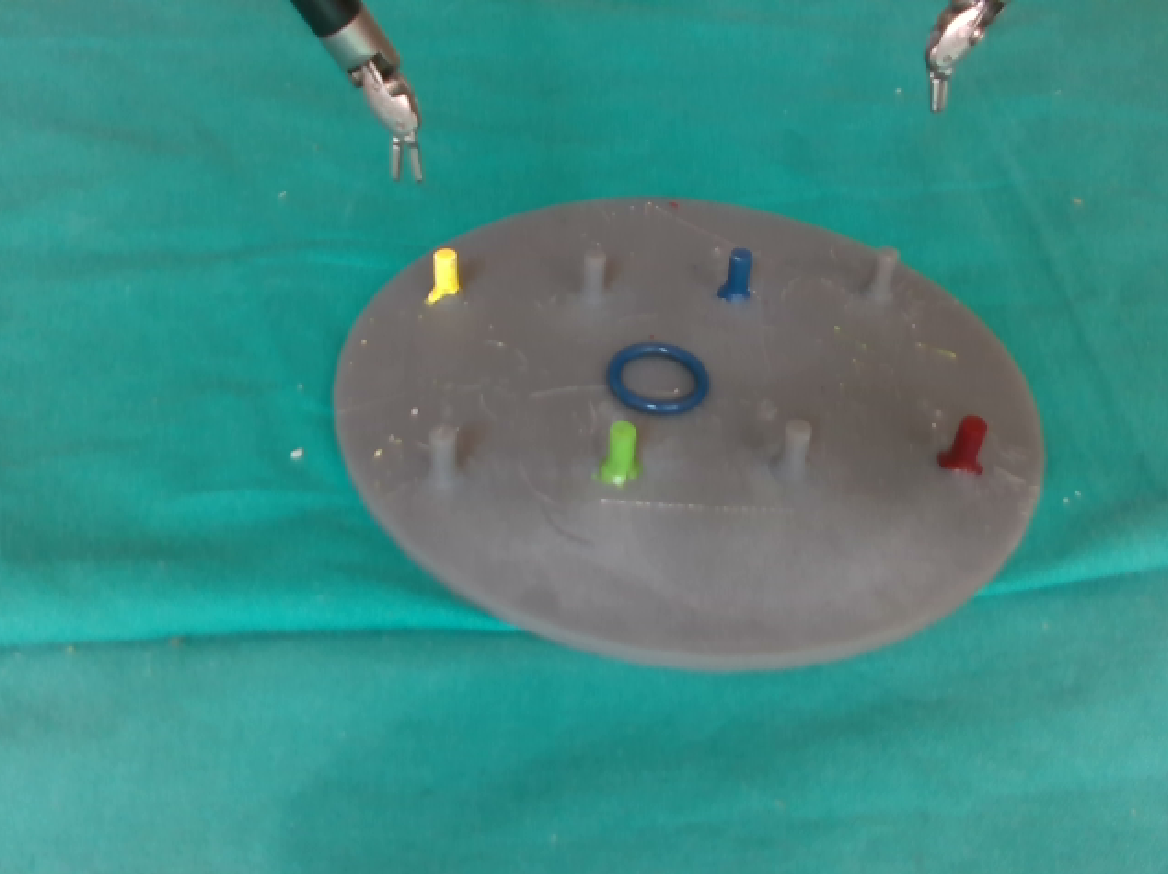}\label{fig:grasp_test}}
    \hfill
    \subfloat[Dual-arm manipulation
    ]{\includegraphics[height=0.35\columnwidth]{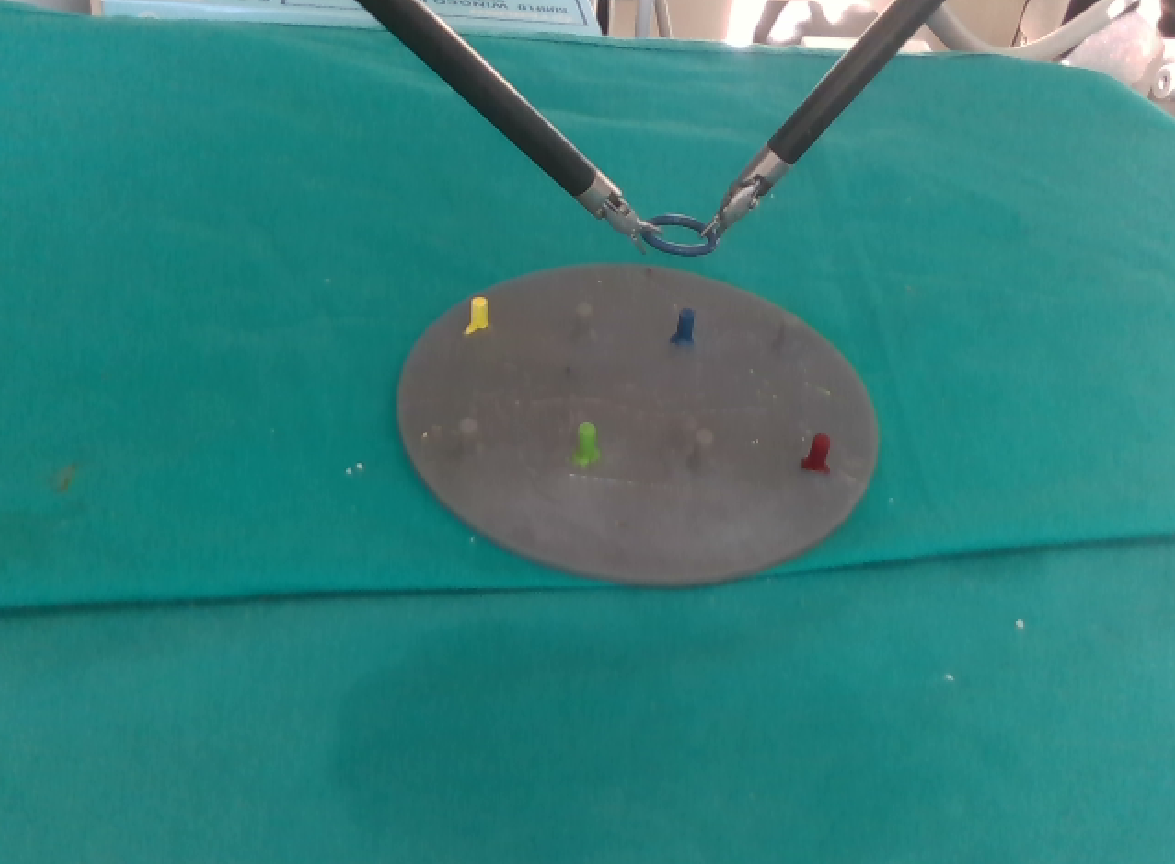}\label{fig:bimanual_test}}
    \caption{Scenarios for validation of the calibration procedure (PSM1 right, PSM2 left).}
    \label{fig:scenes}
\end{figure}

After this preliminary test, we validate our calibration method in two benchmark scenarios for surgical robotics shown in Figure \ref{fig:scenes}: (A) localization and grasping, and (B) dual-arm manipulation.

\subsection{Localization and grasping}
In the first scenario (Figure \ref{fig:grasp_test}) the two PSMs must autonomously grasp the blue ring on the peg base in two different points. The RGB-D camera identifies the point cloud corresponding to the ring after color and shape segmentation, and points are transformed from the camera to the common reference frame. The ring has a diameter of 1.5 \si{\cm}, and the target points for PSM1 and PSM2 are determined as the rightmost and the leftmost in the point cloud, respectively. We place the ring in different positions on the peg base (including the center of the base, two positions on both sides, and the four corners of the base in Figure \ref{fig:grasp_test}) and we compute the Euclidean distance between the grasping points estimated from the camera with the points reached by the PSMs. In this way, we estimate the sum of the precision of our calibration procedure on the $x-y$ plane and the intrinsic kinematic precision of the \davinci. The estimated kinematic precision of the robot is 1.02 \si{\mm} on average (0.58 \si{\mm} standard deviation) when localizing and reaching fiducial markers \cite{haidegger2010importance}, with a maximum error of 2.72 \si{\mm} \cite{kwartowitz2006toward}. The results of our first test are reported in Table \ref{table:errors_grasp}.

\begin{table}[h]
\centering
    \caption{Errors between grasping points reached by the PSMs and the camera estimation for test in Figure \ref{fig:grasp_test}. Positions of PSMs are referred to the common reference frame. Average calibration error is 1.65 \si{\mm}.}
\label{table:errors_grasp}

\begin{tabular}{cccc}
PSM1 (x,y,z) & PSM1 error & PSM2 (x,y,z) & PSM2 error  \\ 
(\si{\mm}) & (\si{\mm}) & (\si{\mm}) & (\si{\mm}) \\
\hline \\
(5.5, 1.0, 23.1) & 1.890 & (3.6, 18.4, -21.7) & 1.125 \\
(11.7, 10.6, -23.2) & 1.221 & (7.2, 30.4, -1.6) & 0.704 \\
(6.5, 28.6, -22.2) & 0.986 & (7.9, 48.2, -21.9) & 1.267 \\
(14.0, -31.2, -24.5) & 1.970 &  (9.7, -11.8, -24.0) & 3.356 \\
(19.7, -52.6, -26.3) & 1.905 & (14.3, -32.7, -25.5) & 1.760 \\
(72.3, -7.2, -28.2) & 2.058 & (77.2, 12.3, -28.5) & 4.745 \\
(-8.8, -57.3, -25.8) & 1.550 & (-12.9, -35.7, -23.5) & 0.621 \\
(-15.3, -3.4, -23.4) & 0.902 & (-17.7,  16.1, -22.1) & 2.141 \\
(106.4, -69.1, -37.6) & 4.727 &  (79.1, -47.2, -27.7) & 2.058 \\
\end{tabular}
\end{table}

We notice that the error has no neat dependency on the positioning on the plane. The overall average error is 1.94 \si{\mm} (1.21 \si{\mm} standard deviation), with a maximum overall error of 4.75 \si{\mm}. Assuming the intrinsic kinematic error is independent on the calibration error, the precision of our calibration procedure on the plane is approximately 1.65 \si{\mm}, less than one half of the state of the art reached in \cite{table, realsense_kuka}.

\subsection{Dual arm manipulation}
\begin{figure*}[t]
 \centering
     \subfloat[$r = 1 cm$]{\includegraphics[height=0.5\columnwidth]{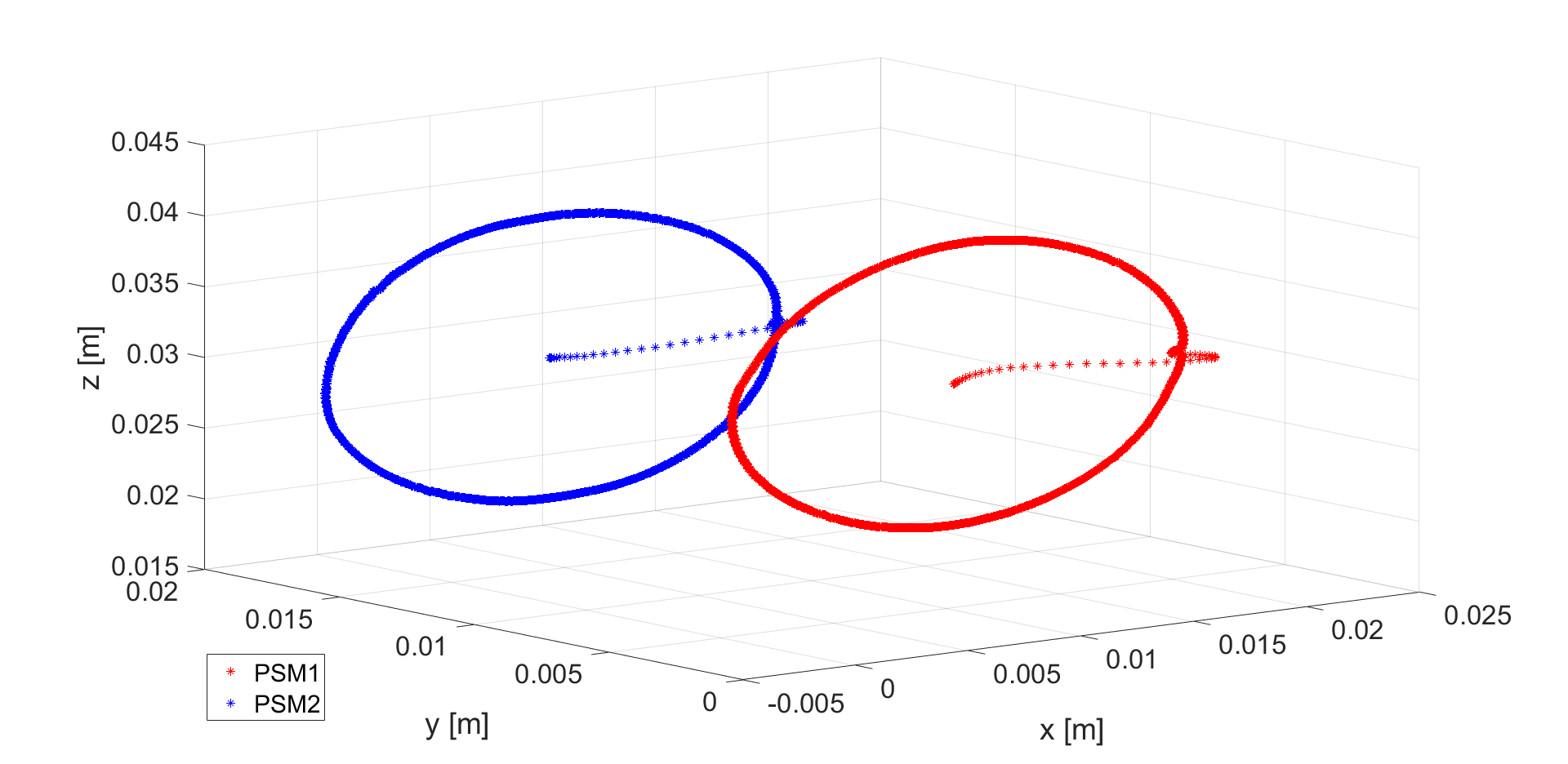}\label{fig:rad1}}
     \subfloat[$r = 5 cm$
     ]{\includegraphics[height=0.5\columnwidth]{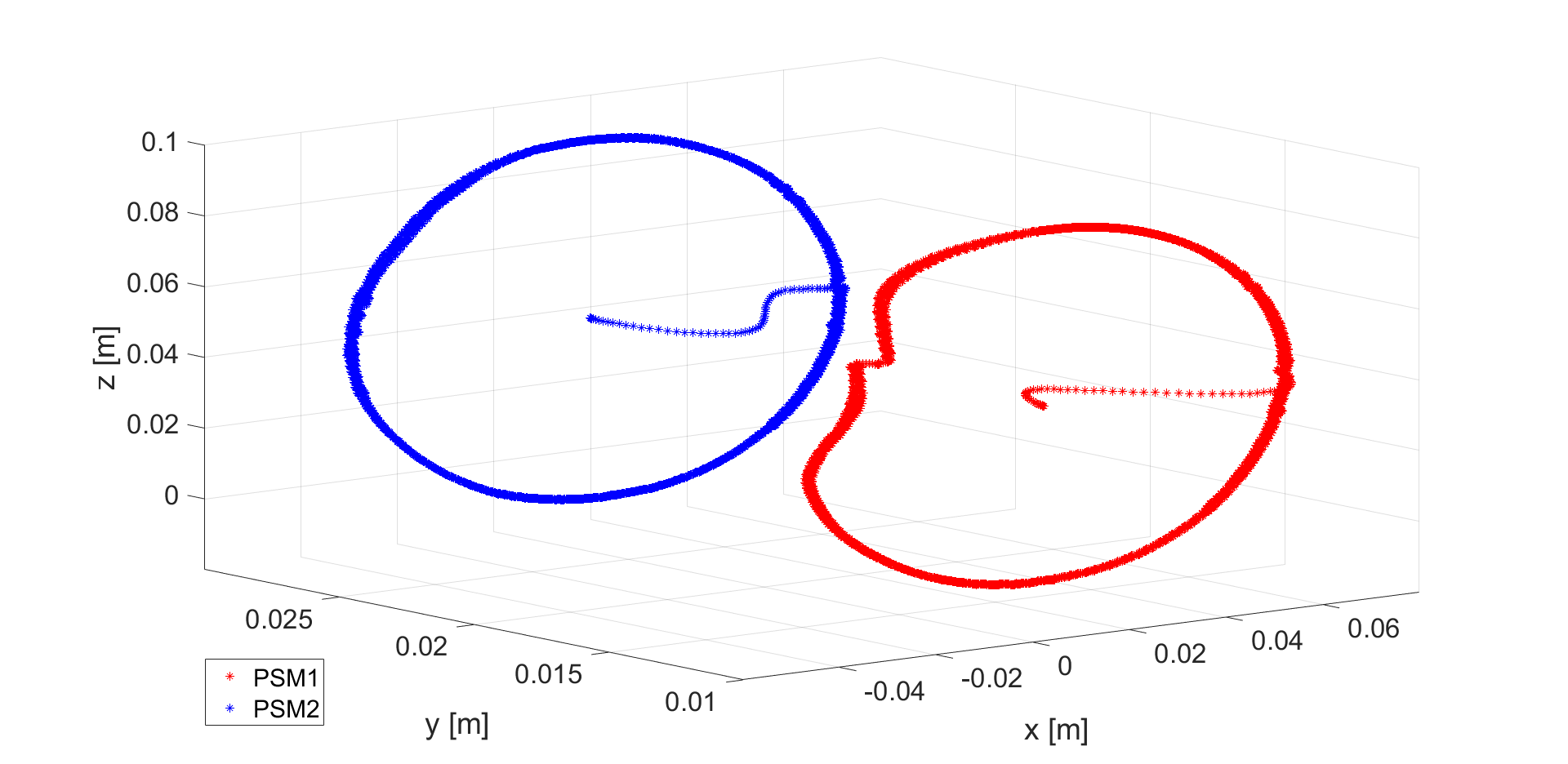}\label{fig:rad5}}
     \caption{Trajectories executed by the PSMs when holding the same ring (in common reference frame) for test in Figure \ref{fig:bimanual_test}.}
\label{fig:circles}
\end{figure*}

In the second scenario (Figure \ref{fig:bimanual_test}) the PSMs start holding the same ring, and they must execute simultaneous circular trajectories in the $x-z$ plane with radius $r$ ranging from 1 to 5 \si{\cm}. The trajectories are pre-computed in the common reference frame, and the PSMs are commanded with the transformed waypoints in their relative frames. This task validates the accuracy of the transformations between the arms resulting from our calibration procedure. Trajectories executed from the PSMs are shown in Figure \ref{fig:circles}. All trajectories start from the same initial position in order to make the analysis of performances more robust. Notice that for $r = 5 \si{\cm}$ the PSMs reach the limit of the workspace and deformation in the trajectory becomes visible. Table \ref{table:results_manip} shows quantitative results for this test. We measure the difference between the trajectories of the two PSMs, and we consider the standard deviation and the maximum deviation from the mean for each radius. In absence of calibration and kinematic errors, the difference between the trajectories would have null standard deviation. The variability on the difference between the trajectories increases with the radius of the circumference, hence with the $z$ coordinate and as the arms approach the limits of their workspace (as already studied in \cite{kwartowitz2006toward}). In the worst case scenario ($r = 5 \si{\cm}$) the overall standard deviation is 1.71 \si{\mm}. Removing the intrinsic precision of the \davinci leads to a (maximum) average calibration error of 1.38 \si{\mm}, confirming the improvement shown in our first test with respect to the state of the art. This proves the comparable precision of our calibration method along all directions in the Euclidean space.


\begin{table}[t]
\centering
\caption{Similarity between the trajectories executed by the PSMs holding the ring for test in Figure \ref{fig:bimanual_test}.}
\label{table:results_manip}
\begin{tabular}{ccc}
Radius & Standard deviation & Max deviation \\
(\si{\mm}) & (\si{\mm}) & (\si{\mm}) \\
\hline \\
10 & 0.23               & 1.89          \\
20 & 0.42               & 3.74          \\
30 & 0.66               & 6.16          \\
40 & 0.85               & 8.50          \\
50 & 1.71               & 7.33         
\end{tabular}
\end{table}

\section{Conclusion}\label{sec:conclusion}
In this paper we proposed a novel 3D calibration procedure for the slave manipulators and the ECM of the \davinci surgical robot. Our procedure exploits a RGB-D Realsense camera. We have validated our calibration procedure on two relevant use cases for surgery, localization and grasping of a small object and dual-arm manipulation. Both tasks require an accurate estimation of the transformation tree connecting the arms and the camera, to guarantee precise positioning and coordination of the PSMs. We have compared the errors measured in the two scenarios with the intrinsic kinematic precision of the \davinci, to estimate the precision of our calibration procedure only. The results show that our method outperforms the state-of-the-art solutions with RGB-D camera, and guarantees sufficient precision for executing surgical tasks (1.65 \si{\mm} in the calibration plane and 1.38 \si{\mm} in the normal plane).

The main drawback of our solution is the use of a RGB-D camera, which limits its application in real surgery.
However, our methodology can be easily extended to a setup with a standard surgical endoscope. 
The main issue with an endoscope is that the small baseline between the stereo cameras introduces additional complexities and reduces the depth range of view. We will address this problem in our future research. Moreover, we will develop an autonomous procedure for our calibration method, which can significantly reduce manual errors and prompt its implementation in real surgical systems.

\section*{Acknowledgment}
This work has received funding from the European Research Council (ERC) under the European Unions Horizon 2020 research and innovation programme under grant agreement No. 742671 (ARS) and No. 779813 (SARAS).
\ifCLASSOPTIONcaptionsoff
  \newpage
\fi

\bibliographystyle{IEEEtran}
\bibliography{IEEEabrv,biblio.bib}

\begin{thebibliography}{10}
\providecommand{\url}[1]{#1}
\csname url@samestyle\endcsname
\providecommand{\newblock}{\relax}
\providecommand{\bibinfo}[2]{#2}
\providecommand{\BIBentrySTDinterwordspacing}{\spaceskip=0pt\relax}
\providecommand{\BIBentryALTinterwordstretchfactor}{4}
\providecommand{\BIBentryALTinterwordspacing}{\spaceskip=\fontdimen2\font plus
\BIBentryALTinterwordstretchfactor\fontdimen3\font minus
  \fontdimen4\font\relax}
\providecommand{\BIBforeignlanguage}[2]{{%
\expandafter\ifx\csname l@#1\endcsname\relax
\typeout{** WARNING: IEEEtran.bst: No hyphenation pattern has been}%
\typeout{** loaded for the language `#1'. Using the pattern for}%
\typeout{** the default language instead.}%
\else
\language=\csname l@#1\endcsname
\fi
#2}}
\providecommand{\BIBdecl}{\relax}
\BIBdecl

\bibitem{du2016combined}
X.~Du, M.~Allan, A.~Dore, S.~Ourselin, D.~Hawkes, J.~D. Kelly, and D.~Stoyanov,
  ``Combined 2d and 3d tracking of surgical instruments for minimally invasive
  and robotic-assisted surgery,'' \emph{International journal of computer
  assisted radiology and surgery}, vol.~11, no.~6, pp. 1109--1119, 2016.

\bibitem{ARS2019}
P.~Fiorini, D.~Dall'Alba, M.~Ginesi, B.~Maris, D.~Meli, H.~Nakawala,
  A.~Roberti, and E.~Tagliabue, ``Challenges of autonomous robotic surgery,''
  in \emph{Hamlyn Symposium on Medical Robotics (HSMR)}, 2019.

\bibitem{DeRossi2019a}
G.~De~Rossi, M.~Minelli, A.~Sozzi, N.~Piccinelli, F.~Ferraguti, F.~Setti,
  M.~Bonf{\'e}, C.~Secchi, and R.~Muradore, ``Cognitive robotic architecture
  for semi-autonomous execution of manipulation tasks in a surgical
  environment,'' in \emph{2019 IEEE/RSJ International Conference on Intelligent
  Robots and Systems (IROS)}.\hskip 1em plus 0.5em minus 0.4em\relax IEEE,
  2019, pp. 7827--7833.

\bibitem{adebar20143}
T.~K. Adebar, A.~E. Fletcher, and A.~M. Okamura, ``3-d ultrasound-guided
  robotic needle steering in biological tissue,'' \emph{IEEE Transactions on
  Biomedical Engineering}, vol.~61, no.~12, pp. 2899--2910, 2014.

\bibitem{ISMR2020_altair}
N.~Piccinelli, A.~Roberti, E.~Tagliabue, F.~Setti, G.~Kronreif, R.~Muradore,
  and P.~Fiorini, ``Rigid 3d registration of pre-operative information for
  semi-autonomous surgery,'' in \emph{2020 International Symposium on Medical
  Robotics (2020) - Atlanta (USA)}, I.~S. on~Medical~Robotics, Ed.\hskip 1em
  plus 0.5em minus 0.4em\relax International Symposium on Medical Robotics,
  2020.

\bibitem{table}
S.~Kahn, D.~Haumann, and V.~Willert, ``Hand-eye calibration with a depth
  camera: 2d or 3d?'' in \emph{2014 International Conference on Computer Vision
  Theory and Applications (VISAPP)}, vol.~3.\hskip 1em plus 0.5em minus
  0.4em\relax IEEE, 2014, pp. 481--489.

\bibitem{shah2012overview}
M.~Shah, R.~D. Eastman, and T.~Hong, ``An overview of robot-sensor calibration
  methods for evaluation of perception systems,'' in \emph{Proceedings of the
  Workshop on Performance Metrics for Intelligent Systems}, 2012, pp. 15--20.

\bibitem{shiu1989calibration}
Y.~C. Shiu and S.~Ahmad, ``Calibration of wrist-mounted robotic sensors by
  solving homogeneous transform equations of the form ax= xb.'' \emph{IEEE
  Transactions on robotics and automation}, vol.~5, no.~1, pp. 16--29, 1989.

\bibitem{chou1991finding}
J.~C. Chou and M.~Kamel, ``Finding the position and orientation of a sensor on
  a robot manipulator using quaternions,'' \emph{The international journal of
  robotics research}, vol.~10, no.~3, pp. 240--254, 1991.

\bibitem{liang2008hand}
R.-h. Liang and J.-f. Mao, ``Hand-eye calibration with a new linear
  decomposition algorithm,'' \emph{Journal of Zhejiang University-SCIENCE A},
  vol.~9, no.~10, pp. 1363--1368, 2008.

\bibitem{pan2014closed}
H.~Pan, N.~L. Wang, and Y.~S. Qin, ``A closed-form solution to eye-to-hand
  calibration towards visual grasping,'' \emph{Industrial Robot: An
  International Journal}, 2014.

\bibitem{najafi2014closed}
M.~Najafi, N.~Afsham, P.~Abolmaesumi, and R.~Rohling, ``A closed-form
  differential formulation for ultrasound spatial calibration: multi-wedge
  phantom,'' \emph{Ultrasound in medicine \& biology}, vol.~40, no.~9, pp.
  2231--2243, 2014.

\bibitem{kim2013hand}
D.~W. Kim and J.~E. Ha, ``Hand/eye calibration using 3d-3d correspondences,''
  in \emph{Applied Mechanics and Materials}, vol. 319.\hskip 1em plus 0.5em
  minus 0.4em\relax Trans Tech Publ, 2013, pp. 532--535.

\bibitem{fuchs2012calibration}
S.~Fuchs, ``Calibration and multipath mitigation for increased accuracy of
  time-of-flight camera measurements in robotic applications,'' Ph.D.
  dissertation, Universit{\"a}tsbibliothek der Technischen Universit{\"a}t
  Berlin, 2012.

\bibitem{fischler1981random}
M.~A. Fischler and R.~C. Bolles, ``Random sample consensus: a paradigm for
  model fitting with applications to image analysis and automated
  cartography,'' \emph{Communications of the ACM}, vol.~24, no.~6, pp.
  381--395, 1981.

\bibitem{realsense_kuka}
S.~Qiu, M.~Wang, and M.~R. Kermani, ``A new formulation for hand-eye
  calibrations as point set matching,'' \emph{IEEE Transactions on
  Instrumentation and Measurement}, 2020.

\bibitem{haidegger2010importance}
T.~Haidegger, P.~Kazanzides, I.~Rudas, B.~Beny{\'o}, and Z.~Beny{\'o}, ``The
  importance of accuracy measurement standards for computer-integrated
  interventional systems,'' in \emph{EURON GEM Sig Workshop on The Role of
  Experiments in Robotics Research at IEEE ICRA}, 2010, pp. 1--6.

\bibitem{kwartowitz2006toward}
D.~M. Kwartowitz, S.~D. Herrell, and R.~L. Galloway, ``Toward image-guided
  robotic surgery: determining intrinsic accuracy of the da vinci robot,''
  \emph{International Journal of Computer Assisted Radiology and Surgery},
  vol.~1, no.~3, pp. 157--165, 2006.

\end{thebibliography}








\end{document}